%% file: paper.tex
\documentclass[
]{ceurart}

\usepackage{listings}
\usepackage{booktabs}% for nice tables
\usepackage{amsthm}% nice theorem formatting
\usepackage{amsmath}% improve math presentation
\usepackage{amsfonts}
\usepackage{mathtools}
\usepackage{graphicx}% used for includegraphics
\usepackage{subfigure}
\usepackage{algorithm}

\usepackage{makecell}

\usepackage{float}% for figure placement

\usepackage[noend]{algpseudocode}

\usepackage{url}

\input{macros}

\begin{document}

\copyrightyear{2024}
\copyrightclause{Copyright for this paper by its authors.
  Use permitted under Creative Commons License Attribution 4.0
  International (CC BY 4.0).}

\conference{HI-AI@KDD, Human-Interpretable AI Workshop at the KDD 2024, 26$^{th}$ of August 2024, Barcelona, Spain }

\title{GLEAMS: Bridging the Gap Between Local and Global Explanations}

\author[1]{Giorgio Visani}[
email=giorgio.visani@leitha.eu]
\cormark[1]
 \address[1]{Leithà S.r.l. - Unipol Group, via Stalingrado 53, Bologna, Italy}

 \author[2]{Vincenzo Stanzione}[%
     email=v.m.stanzione@gmail.com
 ]
 \address[2]{Computer Science Department, Bologna University}

 \author[3]{Damien Garreau}[
     email=damien.garreau@uni-wuerzburg.de]
     %\fnmark[1]
\address[3]{Julius-Maximilians-Universit\"at W\"urzburg, Germany}

%% Footnotes
\cortext[1]{Corresponding author.}
%\fntext[1]{These authors contributed equally.}

%%
%% The abstract is a short summary of the work to be presented in the
%% article.
\begin{abstract}
The explainability of machine learning algorithms is crucial, and numerous methods have emerged recently. Local, post-hoc methods assign an attribution score to each feature, indicating its importance for the prediction. However, these methods require recalculating explanations for each example. On the other side, while there exist global approaches they often produce explanations that are either overly simplistic and unreliable or excessively complex. To bridge this gap, we propose GLEAMS, a novel method that partitions the input space and learns an interpretable model within each sub-region, thereby providing both faithful local and global surrogates. We demonstrate GLEAMS' effectiveness on both synthetic and real-world data, highlighting its desirable properties and human-understandable insights.
\end{abstract}

%%
%% Keywords. The author(s) should pick words that accurately describe
%% the work being presented. Separate the keywords with commas.

\maketitle

%%%%%%%%%%%%%%%%%%%%%%%%%%%%%%%%%%%%%%%%%%%%%%%%%%%%%%%%%%%%%%%%%%%%%%

\section{Introduction}\label{sec:intro}

In numerous applications, the data present themselves as large arrays---tabular data, where each line corresponds to a data point and each column to a feature. 
In this setting, deep learning is becoming as popular as it already is for more structured data types, and deep neural networks achieve state-of-the-art in many scenarios
\cite{borisov_et_al_2021}. 
A caveat of this approach is the difficulty to obtain insights on a specific prediction: the intricate architectures and the sheer number of parameters prevent the user of getting a clear picture of what is actually important to the model. 

Since getting such \emph{explanations} is desirable in many applications, and could even become mandated by law in certain regions \cite{wachter_et_al_2017}, many interpretability methods have been proposed. 
Without getting into details (we refer to the recent surveys \cite{adadi_berrada_2018} and \cite{linardatos_et_al_2020} for this purpose), we want to make two distinctions clear. 
First, it is challenging for interpretable models to have the same accuracy as non-interpretable ones. 
As a consequence, many explainability methods rely on  a \emph{post hoc} approach, \emph{i.e.}, producing explanations in a second step, exploiting already trained black-box models. 
This is the road we take, since \emph{post-hoc} methods allow us to deal with any given model and not being dependent on a specific architecture. 
Second, the explanations can either be \emph{local} (related to a specific example), or \emph{global} (for all the input space). 
A problem with local explanations is that they have to be computed individually for each new example, which can be time-consuming. 
For instance, LIME \cite{ribeiro_et_al_2016} generates $5,000$ new datapoints for each example to explain. 
The black-box model then has to be queried on each of these points.

In this paper, we propose \textbf{GLEAMS} (Global \& Local ExplainAbility of black-box Models through Space partitioning), a \emph{post hoc} interpretability method providing both global and local explanations. 
Our goal is to build a \emph{global surrogate}~$\surrogatemodel$ of~$\Model$ that shares the global shape of the black-box model~$\Model$ but, at the same time, has a simple local structure. 
Our main motivation in doing so is to extract simultaneously from~$\surrogatemodel$: i) local explanations for any instance; ii) overall importance of the features. 
Secondary objectives are the production of ``what-if''-type explanations, and visual summaries. 
In more details, the core idea of GLEAMS is to recursively split the input space in rectangular cells on which the black-box model can be well-approximated by linear models. 
We describe the method in Section~\ref{sec:method-description}.
The local explanations are then simply given by the feature coefficients of these local surrogate models, while global indicators can be readily computed from those, without querying~$\Model$ model further. 
We describe how to obtain the other indicators in Section~\ref{sec:gleams}, and show experimentally that GLEAMS compares favorably with existing methods in Section~\ref{sec:experiments}. 
GLEAMS code as well as the experiments for this paper are publicly available.\footnote{\url{https://github.com/giorgiovisani/GLEAMS}}
Related work can be found in Appendix~\ref{sec:related}. 

\begin{figure*}[ht!]
\centering
\includegraphics[scale=0.42]{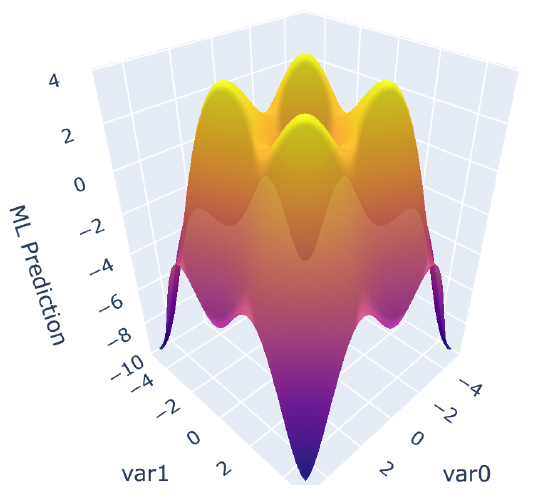}
\hfill 
\includegraphics[scale=0.42]{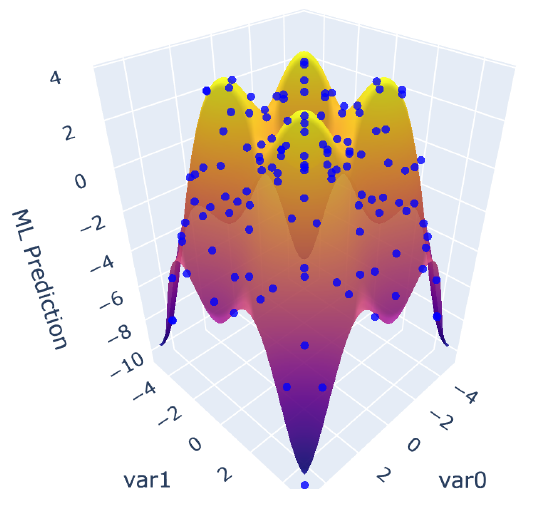}
\hfill 
\includegraphics[scale=0.42]{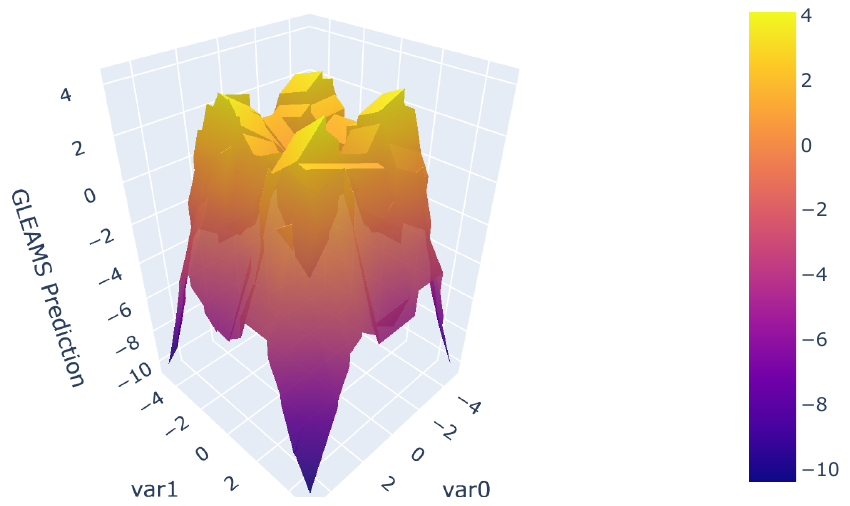}
\hfill 
 \caption{\label{fig:overview}Overview of the global surrogate model construction. \emph{Left panel:} the black-box model maps the input space (here $\inputspace = [0,1]^2$) to $\Reals$, which we can visualize as a surface. \emph{Middle panel:} we generate~$N$ Sobol points on $\inputspace$, giving rise to $N$ measurement points on the surface (in blue). \emph{Right panel:} we fit a piecewise-linear global surrogate model $\surrogatemodel$ on the measurement points by recursively splitting $\inputspace$.}
\end{figure*}

%%%%%%%%%%%%%%%%%%%%%%%%%%%%%%%%%%%%%%%%%

\section{Description of the method}
\label{sec:method-description}

We consider the following general setting: $\Model$ is a mapping from the input space $\inputspace$ to the output space $\labels$. 
This corresponds to the usual situation in machine learning, where $y=\model{x}$ is a good prediction of the true label of $x$. 
In this paper, we assume that $\labels\subseteq \Reals$. 
We want to stress that this can correspond both to the \emph{classification} and \emph{regression} setting. 
In the regression setting, for any input $x\in\inputspace$, $\model{x}$ is simply a real-valued quantity of interest, whereas in the classification case $\model{x}$ corresponds to the pseudo-probability to fall into a given class. 

The construction of the global surrogate on which GLEAMS relies to provide explanations is a two-step process. 
The first step is to sample $N$ points $x_1,\ldots,x_N$, evenly spread in $\inputspace$. 
For each of these points, we query $\Model$ to obtain $y_i=\model{x_i}$. 
Thus we effectively create a dataset $(x_i,y_i)$, which we call \emph{measurement points}. 

Subsequently, the idea is to recursively split the input space~$\inputspace$ and to approximate the model~$\Model$ by a \emph{linear model} on each cell of the partition 
until the linear model in each rectangular leaf fits the measurement points with a sufficiently good accuracy. 
This process is summarized in Figure~\ref{fig:overview}. 
Once this global surrogate constructed, GLEAMS can provide both local and global explanations for any new example to explain. 
Intuitively, given a partition with not too many cells, it is convenient for the user to look at individual features and to visualize globally the model as a piecewise-linear function. 
Concurrently, any new example to explain will fall into a cell, and the user can get a local explanation solely by looking at the coefficients of the local linear model. 
We now give more details on each of these steps.

%%%%%%%%%%%%%%%%%%%%%%%%%%%%%%%%%%%%%%%%%%%%%%%%%%%%%%%%%%%%%%%%%%%%%%%%

\subsection{Measurement points}
\label{sec:measurement-points}

Our main assumption, going into the description of the method, is the following: 

\begin{assumption}
\label{ass:rectangle}
The input space $\inputspace$ is a hyper-rectangle of $\Reals^d$. 
That is, there exist reals numbers $a_j,b_j\in\Reals$ for any $j\in [d]$ such that $\inputspace = \prod_{j=1}^d [a_j,b_j]$. 
\end{assumption}

Intuitively, this corresponds to a situation where each variable has a prescribed range (for instance, age takes values in $[0,120]$). 
In particular, we assume that further examples to explain all fall within this range. 
If they do not, we attribute to these points the explanations corresponding to the projection of these points on $\inputspace$. 
Note that we do not assume normalized data: $a_j$ and $b_j$ can be arbitrary. 
In practice, the boundaries are either inferred from a training set, or directly provided by the user.

A consequence of Assumption~\ref{ass:rectangle} is that we can easily find a set of points covering $\inputspace$ efficiently. 
To achieve this, we use \textbf{Sobol sequences} \citep{sobol_1985}. 
In a nutshell, a Sobol sequence is a quasi-random sequence of points $x_1,\ldots,x_N$ filling the unit cube $[0,1]^d$ as $N$ grows. 
We use Sobol sequences because (a) we want to be able to provide explanations for the whole input space $\inputspace$, thus the measurement points should cover $\inputspace$ as $N$ grows; and (b) the decision surface of the black-box model can be quite irregular, thus we want low discrepancy. 

%%%%%%%%%%%%%%%%%%%%%%%%%%%%%%%%%%%%%%%%%%%%%%%%%%%%%%%%%%%%%%%%%%%%%%%

\subsection{Splitting criterion}
\label{sec:splitting}

A very convenient way to partition the input space is to recursively split $\inputspace$, cutting a cell if some numerical criterion is satisfied. 
Additionally, we consider splits that are parallel to the axes.  
The main reason for doing so is the \emph{tree structure} that emerges, allowing later on for fast query time of new inputs (linear in the depth of the final tree). 
In this section, we describe the criterion used by GLEAMS. 

A popular approach is to fit a constant model on each cell \citep[CART]{breiman_et_al_1984}. 
We see two problems with constant fits.
First, this tends to produce large and thus hard to interpret trees (see, \emph{e.g.}, \citet{chan_loh_2004}). 
Second, the local interpretability would be lost, since the local models would not depend on the features in that case. 
Thus we prefer to fit linear models on each leaf, and our numerical criterion should indicate, on any given leaf, if such simple model is a good fit for~$\Model$. 
GLEAMS relies on a variant of model-based recursive partitioning \citep[MOB]{zeileis_et_al_2008} based on score computations, described in detail in Appendix~\ref{sec:splitting}.

%%%%%%%%%%%%%%%%%%%%%%%%%%%%%%%%%%%%%%%%%%%%%%%%%%%%%%%%%%%%%%%%%%%%%%%%%%%%%%%

\subsection{Global surrogate model}
\label{sec:global-surrogate}

To build the global surrogate model $\surrogatemodel$, starting from the hyper-rectangle $\inputspace$, we \textbf{iteratively split along the best dimension}.  
This process is stopped once one of two criteria is met: i) leaves have $R^2$ higher than $0.95$ (chosen empirically) or ii) number of points in the leaf is less than $\nmin$ threshold, set to $\nmin=\min(20,d+1)$. 
This recursive procedure is described in Algorithm~\ref{algo:tree-construction}, in Appendix \ref{sec:splitting}. 

$\surrogatemodel$ is then achieved by sampling Sobol points on $\inputspace$, computing the value of the black-box model at these points, and then using Algorithm~\ref{algo:tree-construction}, Appendix \ref{sec:splitting}.

%%%%%%%%%%%%%%%%%%%%%%%%%%%%%%%%%%%%%%%%%%%%%%%%%%%%%%%%%%%%%%%%%%%%%%%%%%%%%%%%%

\subsection{GLEAMS}
\label{sec:gleams}

Let us now assume that we computed the global surrogate model~$\surrogatemodel$, that is, a piecewise-linear model based on a partition of $\inputspace$. 
At query time, each new example to explain $\xi\in\inputspace$ is rooted to the corresponding hyper-rectangle $R$ in the tree structure, and thus associated to a linear model $\betahat^\xi$. 
From this local linear model, we can directly get different sorts of explanations. 

\paragraph{Local explanations.}
Simply looking at the coefficient $\betahat^\xi_j$ tells us how important feature $j$ is to predict $\model{\xi}$, locally. 
The intuition here is that $\Model$ is well-approximated by $x\mapsto (\betahat^\xi)^\top x$ on the cell $R$, thus local variations of the model along specific axes are well-described by the coefficients of $\betahat^\xi$. 
These local explanations can be obtained in \emph{constant time}, without re-sampling and querying the model. 
Moreover, we have precise insights regarding the \emph{scale} of these explanations, which is given by the size of~$R$. 
We see this as an advantage of using GLEAMS: the portion of space on which the black-box model is locally approximated is well-identified and accessible to the user. 
This is not the case with other methods such as LIME.

\paragraph{Global explanations.}
We define the global importance of feature~$j$ as the aggregation of the local importance of feature~$j$ on all cells of the partition~$\partition$ of $\inputspace$. 
There are several ways to aggregate those, 
GLEAMS outputs the weighted average of the absolute value of the local importance. 
More precisely, let us define $\betahat^R$ the local model on hyper-rectangle $R\in\partition$, 
 
then the global importance of feature~$j$ is given by 
\begin{equation}
\label{eq:def-global-importance}
I_j \defeq \frac{1}{\volume{\inputspace}}\sum_{R\in\partition} \volume{R} \abs{\betahat^R_j}
\, ,
\end{equation}
where $\volume{E}$ denotes the $d$-dimensional volume of~$E$. 
The motivation for weighting by the cells' volume in Eq.~\eqref{eq:def-global-importance} is to take into account the fact that even though feature~$j$ is very important in a tiny region of the space, if it is not on all other cells then it is not \emph{globally} important. 
We take the absolute value to take into account the possible sign variations: if feature~$j$ is positively important on many cells but negatively important on a similar number of cells, then simply taking the average would tell us that the feature is not globally important.
Such \emph{global attributions} provide a faithful variables ranking, based on the impact they show on~$\Model$ predictions.

%%%%%%%%%%%%%%%%%%%%%%%%%%%%%%%%%%%%%%%%%%%%%%%%%%

\begin{figure*}[t!]
\centering
\includegraphics[scale=0.22]{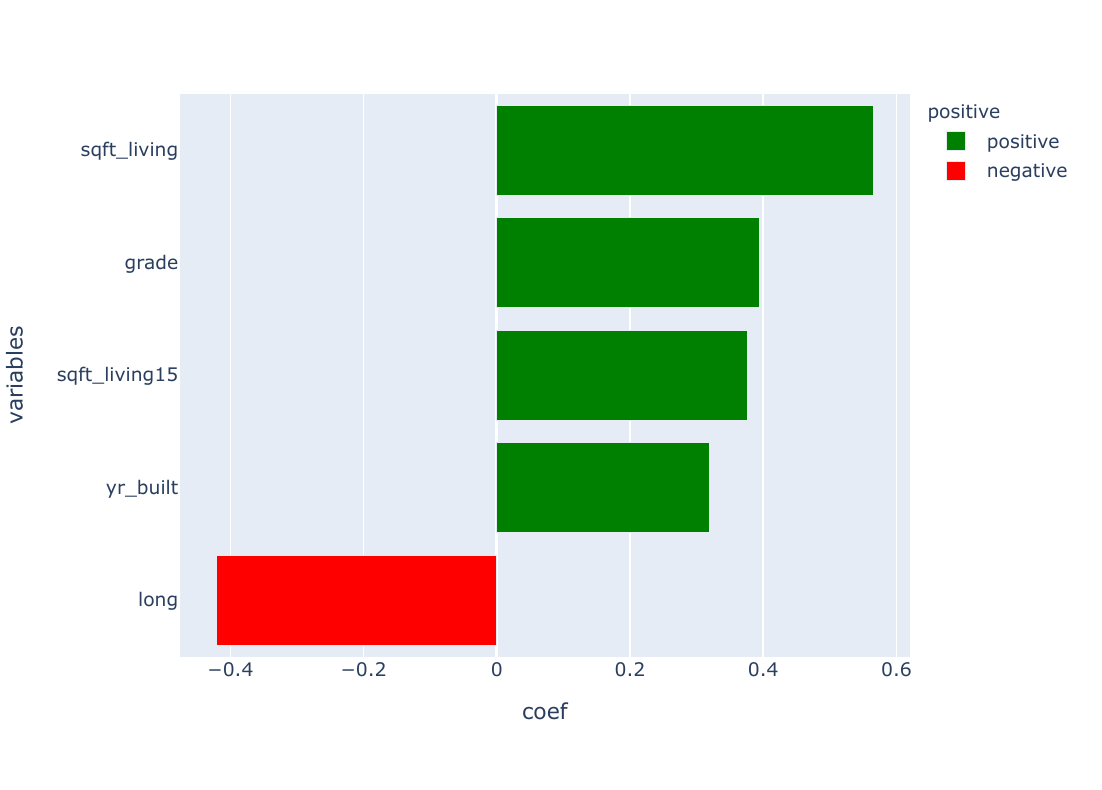}
\hspace{1cm}
\includegraphics[scale=0.31]{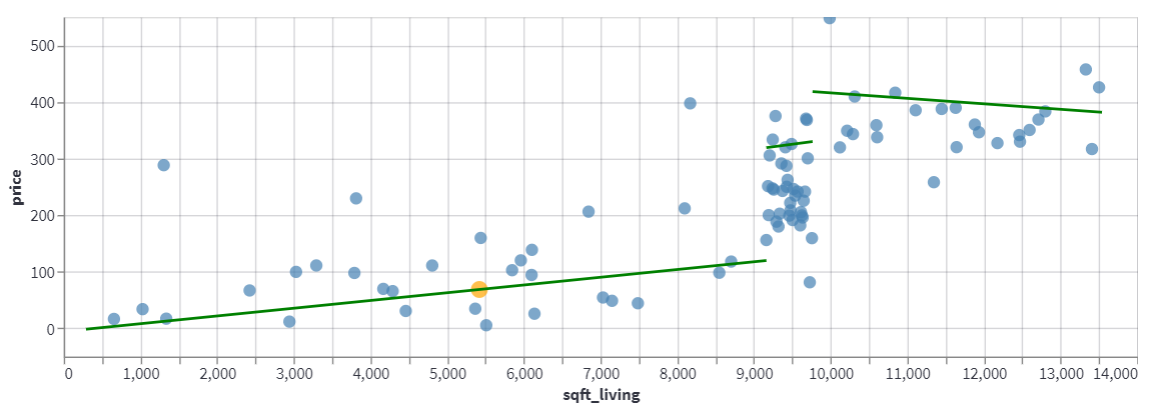}
\caption{\label{fig:gleams-explanations}Gleams explanations consist of both: i) Local importance (feature attribution) and ii) counterfactuals, \emph{i.e.}, answering what-if questions. On the left feature importance for a regressor trained on the house sell dataset (Section~\ref{sec:experiments}), while figure on the right answers to ``What if my house was bigger?'' Moving along feature \texttt{sqft\_living}, for a given example (yellow dot), we can visualize specific values of $\Model$ (Sobol points in blue) as well as the GLEAMS linear approximations (in green).}
\end{figure*}

%%%%%%%%%%%%%%%%%%%%%%%%%%%%%%%%%%%%%%%%%%%%%%%

\paragraph{Counterfactual explanations.}
Querying this new model $\surrogatemodel^\xi_j$ at different values of $x_j$, we can answer to the question ``what would be the decision if we changed feature $j$ by this amount?'' 
More precisely, for a given example $\xi$ and a feature~$j$, we can look at the evolution of $\surrogatemodel$ when all coordinates of the input are fixed to $\xi$, except the $j$th: it suffices to travel in the cells intersecting the line of equation $x_j=\xi_j$ and to read the coefficients of the associated linear models. 
To be precise, we compute 
\begin{equation}
\label{eq:def-what-if}
W_j(x) \defeq \surrogatemodel(\xi_1,\ldots,\xi_{j-1},x,\xi_{j+1},\ldots,\xi_d)
\, .
\end{equation}
This can be computed quickly, since, again, there is no sampling and further calls to $\Model$.  
This allows us to present the user with a cursor which can be moved to set the feature value and present explanations in real time (see Figure~\ref{fig:gleams-explanations}).

%%%%%%%%%%%%%%%%%%%%%%%%%%%%%%%%%%%%%%%%%%%%%%%%%%%%%%%%%%%%%%%%%

\section{Experiments on real data}
\label{sec:experiments}

Our goal here is to show that GLEAMS provides explanations whose quality is comparable to that of other attribution approaches, while providing these attributions in constant time and additional insights. We benchmark the various methods on the \emph{Wine} dataset \citep{cortez_et_al_2009}, \emph{House sell} dataset\footnote{available at \url{https://www.openml.org/search?type=data&status=active&id=42092}} and \emph{Parkinson telemonitoring} dataset \citep{tsanas_et_al_2009}.
We refer to Appendix~\ref{sec:toy-data} for more experiments.

%%%%%%%%%%%%%%%%%%%%%%%%%%%%%%%%%%%%%%%%%%%%%%%%

\begin{table*}\centering
\caption{\label{table:results}Experimental results (higher is better), 5th and 95th confidence intervals are to be found in brackets. GLEAMS is on par with state-of-the-art explanation methods in terms of reliability of explanations.}
\begin{tabular}{@{}lccccccccc@{}}\toprule 
dataset & evaluation & \multicolumn{2}{c}{LIME} & \multicolumn{2}{c}{SHAP} & \multicolumn{2}{c}{PDP} & \multicolumn{2}{c}{\textbf{GLEAMS}} \\
 & & XGB & MLP & XGB & MLP & XGB & MLP & XGB & MLP \\
\midrule 
wine & local & 0.38 & 0.51 & 0.40 & 0.53 & 0.56 & 0.84 & 0.51 & 0.77 \\
 &  & \tiny{[-0.09, 0.77]} & \tiny{[0.19, 0.84]} & \tiny{[-0.08, 0.76]} & \tiny{[0.22, 0.78]} & \tiny{[0.23, 0.83]} & \tiny{[0.69, 0.95]} & \tiny{[0.15, 0.78]} & \tiny{[0.7, 0.86]} \\
wine & global & 0.24 & 0.74 & 0.27 & 0.78 & 0.70 & 0.89 & 0.36 & 0.85 \\
 &  & \tiny{[-0.27, 0.66]} & \tiny{[0.53, 0.9]} & \tiny{[-0.19, 0.65]} & \tiny{[0.53, 0.93]} & \tiny{[0.44, 0.9]} & \tiny{[0.78, 0.97]} & \tiny{[-0.19, 0.79]} & \tiny{[0.69, 0.95]} \\
wine & \makecell{recall \\ ($6$ true features)} & 0.89 & 0.80 & 1.0 & 0.87 & 0.50 & 0.50 & 0.77 & 0.75 \\
 &  & \tiny{[0.67, 1.0]} & \tiny{[0.67, 1.0]} & \tiny{[1.0, 1.0]} & \tiny{[0.83, 1.0]} & \tiny{[0.5, 0.5]} & \tiny{[0.5, 0.5]} & \tiny{[0.5, 0.83]} & \tiny{[0.5, 1.0]} \\
\midrule 
house & local & 0.05 & 0.63 & 0.52 & 0.82 & 0.51 & 0.83 & 0.37 & -0.19 \\
 &  & \tiny{[-0.25, 0.32]} & \tiny{[0.37, 0.81]} & \tiny{[0.22, 0.79]} & \tiny{[0.73, 0.89]} & \tiny{[0.73, 0.9]} & \tiny{[0.74, 0.9]} & \tiny{[0.26, 0.47]} & \tiny{[-0.3, -0.05]} \\
house & global  & 0.23 & 0.63 & 0.37 & 0.81 & 0.38 & 0.82 & 0.47 & 0.24 \\
 &  & \tiny{[-0.09, 0.55]} & \tiny{[0.36, 0.1]} & \tiny{[0.07, 0.64]} & \tiny{[0.69, 0.87]} & \tiny{[0.26, 0.54]} & \tiny{[0.71, 0.9]} & \tiny{[0.07, 0.72]} & \tiny{[-0.02, 0.57]} \\
house & \makecell{recall \\($10$ true features)} & 0.85 & 0.74 & 0.99 & 0.69 & 0.40 & 0.40 & 0.61 & 0.58 \\
 &  & \tiny{[0.7, 1.0]} & \tiny{[0.6, 0.8]} & \tiny{[0.9, 1.0]} & \tiny{[0.5, 0.8]} & \tiny{[0.4, 0.4]} & \tiny{[0.4,0.4]} & \tiny{[0.4, 0.7]} & \tiny{[0.4,0.7]} \\
\midrule
Parkinson & local & 0.10 & 0.26 & 0.47 & 0.55 & 0.28 & 0.49 & 0.50 & 0.01 \\
 &  & \tiny{[-0.11, 0.41]} & \tiny{[-0.04, 0.59]} & \tiny{[0.3, 0.66]} & \tiny{[0.28, 0.76]} & \tiny{[0.16, 0.34]} & \tiny{[0.26, 0.76]} & \tiny{[0.36, 0.6]} & \tiny{[-0.2, 0.23]} \\
Parkinson & global  & 0.31 & 0.46 & 0.41 & 0.71 & 0.52 & 0.66 & 0.30 & 0.36 \\
 &  & \tiny{[0.02, 0.59]} & \tiny{[0.16, 0.71]} & \tiny{[0.08, 0.71]} & \tiny{[0.47, 0.88]} & \tiny{[0.29, 0.71]} & \tiny{[0.43, 0.83]} & \tiny{[-0.06, 0.63]} & \tiny{[0.01, 0.68]} \\
Parkinson & \makecell{recall \\ ($10$ true features)} & 0.86 & 0.77 & 0.97 & 0.80 & 0.40 & 0.40 & 0.50 & 0.60 \\
 &  & \tiny{[0.7, 1.0]} & \tiny{[0.5, 0.8]} & \tiny{[0.9, 1.0]} & \tiny{[0.4, 0.7]} & \tiny{[0.4, 0.4]} & \tiny{[0.4, 0.4]} & \tiny{[0.3, 0.7]} & \tiny{[0.43, 0.6]} \\
\bottomrule
\end{tabular}
\end{table*}

%%%%%%%%%%%%%%%%%%%%%%%%%%%%%%%%%%%%%%%%%%%%%%%

\paragraph{Models.}
We considered two different models, trained similarly across all datasets. 
The first is an \emph{XGBoost} model \citep{chen_guestrin_2016}. 
XGBoost iteratively aggregates base regressors greedily minimizing a proxy loss. 
Following \citet{friedman2001greedy} prescriptions, we employ simple CART trees with maximum depth $2$ as base regressors, learning rate of $0.05$, and early stopping rounds set to $100$. 
The second is a \emph{multi-layer perceptron} (MLP), composed by two hidden dense layers of 264 neurons each, trained by adaptive stochastic gradient descent \citep[ADAM]{kingma_ba_2014} and early stopping on a hold-out validation set. 
Both these models achieve state-of-the-art accuracy in many settings and are not inherently interpretable. 

\paragraph{Methods.}
We compared GLEAMS to three methods, namely LIME for tabular data, SHAP, and PDP. 
We used default parameters, except for the number of feature combinations tested in SHAP for the recall metric, which were shrinked to $500$ to limit computing time.
GLEAMS results have been obtained using $N=15$ (number of Sobol points), which guaranteed a running time in the order of 5 minutes for each of the three datasets at hand.

\paragraph{Evaluation.}
We compare GLEAMS to existing methodology using monotonicity and recall. 
For each interpretability method to evaluate, for each point in the test set, we compute the vectors of attributions $\alpha\in\Reals^d$ and  expected loss $e\in\Reals^d$ restricted to the corresponding feature. 
Monotonicity is then defined as the Spearman rank correlation between~$\abs{\alpha}$ and $e$ \cite{nguyen_martinez_2020}. 
Depending on the region where the $e$ values are computed we obtain \textbf{local monotonicity}, \emph{i.e.}, inside each specific GLEAMS leaf, or \textbf{global monotonicity} when considering the whole input space. Let us now consider a model trained to use only a subset $T$ of the features. We can actually be certain that some feature attributions should be~$0$. 
For any given example $\xi$, we can define $T^\xi$ the set of the top $\abs{T}$ features, ranked by $\abs{\alpha_j}$. 
\textbf{Recall of Important Features} \cite{ribeiro_et_al_2016} is the fraction of $T$ features retrieved in $T^\xi$, averaged among the test units. 
More on this in the Appendix \ref{sec:evaluation}.

\paragraph{Results.}
We present our results in Table~\ref{table:results}, reporting  the average metrics taken on the test set. 
All evaluation metrics are described in Appendix~\ref{sec:evaluation}. 

PDP generally achieves better monotonicity metrics, since it is designed precisely to obtain feature \emph{attributions} related to the feature impact on the predictions, but it performs worse on the recall task.
We notice that GLEAMS shows better results on \emph{local} and \emph{global monotonicity} compared to LIME and SHAP on the wine dataset, while GLEAMS performances start to degrade with an increased number of features. In particular we notice the poor performance on \emph{local monotonicity} of the MLP model for the Parkinson dataset. 
The degradation is due to the higher dimensionality of the two datasets, which makes models surface more complex (MLP in particular). 
Increasing $N$ would likely improve the results, enabling smaller partitions and more accurate local models. Unfortunately, we did not have time to run experiments with higher~$N$.
Regarding the recall metric, SHAP stands as the best in class, while GLEAMS is usually slightly worse but still comparable with LIME, and systematically better than PDP.

%%%%%%%%%%%%%%%%%%%%%%%%%%%%%%%%%%%%%%%%%%%%%%%%%%%%%%%%%%%%%%%%%%%%%%%%%%%

\section{Conclusion}
\label{sec:conclusion}

In this paper, we presented GLEAMS, a new interpretability method that approximates a black-box model by recursively partitioning the input space and fitting linear models on the elements of the partition. 
Both local and global explanations can then be obtained in constant time, for any new example, in sharp contrast with existing \emph{post hoc} methods such as LIME and SHAP. 

The main limitation of the method is the assumption that all features are continuous. 
As future work, we aim to extend GLEAMS to mixed features (both categorical and continuous). 
Another limitation of the method is the initial number of model queries needed. 
One possible way to reduce it would be to evaluate the local regularity of~$\Model$ and to sample less when the model is smooth.

%%
%% Define the bibliography file to be used
\bibliography{biblio}

%%%%%%%%%%%%%%%%%%%%%%%%%%%%%%%%%%%%%%%%%%%%%%%%%%%%%%%%%%%%%%%%%

\newpage 
\appendix

\section{Related work}
\label{sec:related}

The idea of approximating globally a complex model such as a deep neural network by a simpler \emph{surrogate model} relying on a tree-based partition of the space can be dated back at \citet{craven_shavlik_1996}.
The proposed method, TREPAN, approximates the outputs of the network on the training set by a decision tree, choosing the splits using the \emph{gain ratio} criterion \citep{quinlan_1993}. 
More recently, inspired by \citet{gibbons_et_al_2013}, \citet{zhou_hooker_2016}
 proposed to use another split criterion, based upon an asymptotic expansion of the Gini criterion. 
Both these approaches are limited to the classification setting, and also require access to the training data (which is implicitly assumed to have a good covering of the input space).

From the interpretability side, a standard way to explain~$\Model$, which is closely related to our approach, is to compute \emph{attributions} scores for each feature. 
\citet{lei_distribution-free_2018} exclude a specific feature from $\Model$ and define its impact as the deterioration of model predictions. 
On the contrary, Partial Dependence Plots (PDP) developed by \citet{friedman2001greedy} measure the sensitivity of $\Model$ to changes in the specific feature, isolating its effect from the ones of the other features. 
\citet{ribeiro_et_al_2016} propose LIME, which proposes as attribution for a specific example the coefficients of a linear model approximating $\Model$ locally. 
Similarly, \citet{lundberg2017unified} introduce the idea of decomposing $\Model$ predictions into single variables \emph{attributions}, using SHAP. The technique samples combinations of features $S$ and average changes in single instance prediction between the restricted $\Model(S\cup i)$ against $\Model(S)$, obtaining local \emph{attributions} for the  feature $i$. Global SHAP \emph{attributions} arise by averaging local attributions over the entire dataset.

Attribution methods usually have to be recomputed for each new example, a caveat that we propose to overcome with GLEAMS, computing a global surrogate and relying on it to provide explanations. 
We are not the first to propose to provide explanations on a global scale:  
\citet{ribeiro_et_al_2018} propose Anchors, a method extracting local subsets of the features (anchors) that are sufficient to recover the same prediction, while having a good global coverage. 
A caveat of the method is that local explanations are not quantitative since all members of a given anchor are equivalent. 
Additionally, the anchors can have many elements if the example at hand is near the decision boundary. 
Let us also mention \citet{setzu_et_al_2021}, which proposes to \emph{aggregate} local explanations: starting from local decision rules, GlocalX combines them in a hierarchical manner to form a global explanation of the model. 
In the \emph{ad hoc} setting, \citet{harder_et_al_2020} proposes to train an interpretable and differentially private model by using several local linear maps per class: the pseudo-probability of belonging to a given class is the softmax of a mixture of linear models. 
This approach is limited to the classification setting, as with GlocalX and Anchors.

%%%%%%%%%%%%%%%%%%%%%%%%%%%%%%%%%%%%%%%%%%%%%%%%%%%%%%%%%%%%%%%%%%%%%%%

\section{Splitting criterion}
\label{sec:splitting}

Intuitively, any well-behaved function can be locally approximated by a linear component, yielding a statistical model which we now describe. 
On any given leaf, for any given feature~$j$, we can reorder the measurement points such that the points contained in the leaf are $x_1,\ldots,x_n$, with $x_{1,j}\leq \cdots \leq x_{n,j}$. 
To these points correspond model values $y_1,\ldots,y_n$. 
In accordance with the intuition given above, we write $y_i = \beta^\top \xtilde_i + \epsilon_i$ for all $i\in [n]$, where $\beta\in\Reals^{d+1}$ are the coefficients of our local linear model, $\xtilde_i\in\Reals^{d+1}$ is the concatenation of $x_i$ with a leading $1$ to take the intercept into account, and $\epsilon_i$ is an error term. 
Let us now follow \citet{merkle_zeileis_2013} and make an i.i.d. Gaussian assumption on the $\epsilon_i$s, giving rise to the likelihood
\begin{equation}
\label{eq:gaussian-likelihood-initial}
\likelihood{\beta;x_i} = \frac{1}{(2\pi\sigma^2)^{d/2}} \expl{\frac{-1}{2\sigma^2}(y_i - \beta^\top\xtilde_i)^2}
\, ,
\end{equation}
where $\sigma^2 >0$ is the variance of $\epsilon_i$ for all $i\in [n]$. 
Taking the log in Eq.~\eqref{eq:gaussian-likelihood-initial}, the log-likelihood of a single observation is given by
\begin{equation}
\label{eq:log-likelihood-initial}
\ell(\beta;x_i) = \frac{-1}{2}\left\{ \frac{1}{\sigma^2} (y_i - \beta^\top\xtilde_i)^2 + d\log 2\pi\sigma^2 \right\}
\, .
\end{equation}
The individual scores are obtained by taking the partial derivatives of the log-likelihood with respect to the parameters,  that is, for any $i\in [n]$, 
\begin{equation}
\label{eq:scores-initial}
\score{\beta;x_i} = \left( \frac{\partial \ell (\beta;x_i)}{\partial \beta_0},\frac{\partial \ell (\beta;x_i)}{\partial \beta_1},\ldots,\frac{\partial \ell (\beta;x_i)}{\partial \beta_d}\right)^\top
\, .
\end{equation}

\begin{figure}[ht]
    \centering
    \includegraphics[scale=0.3]{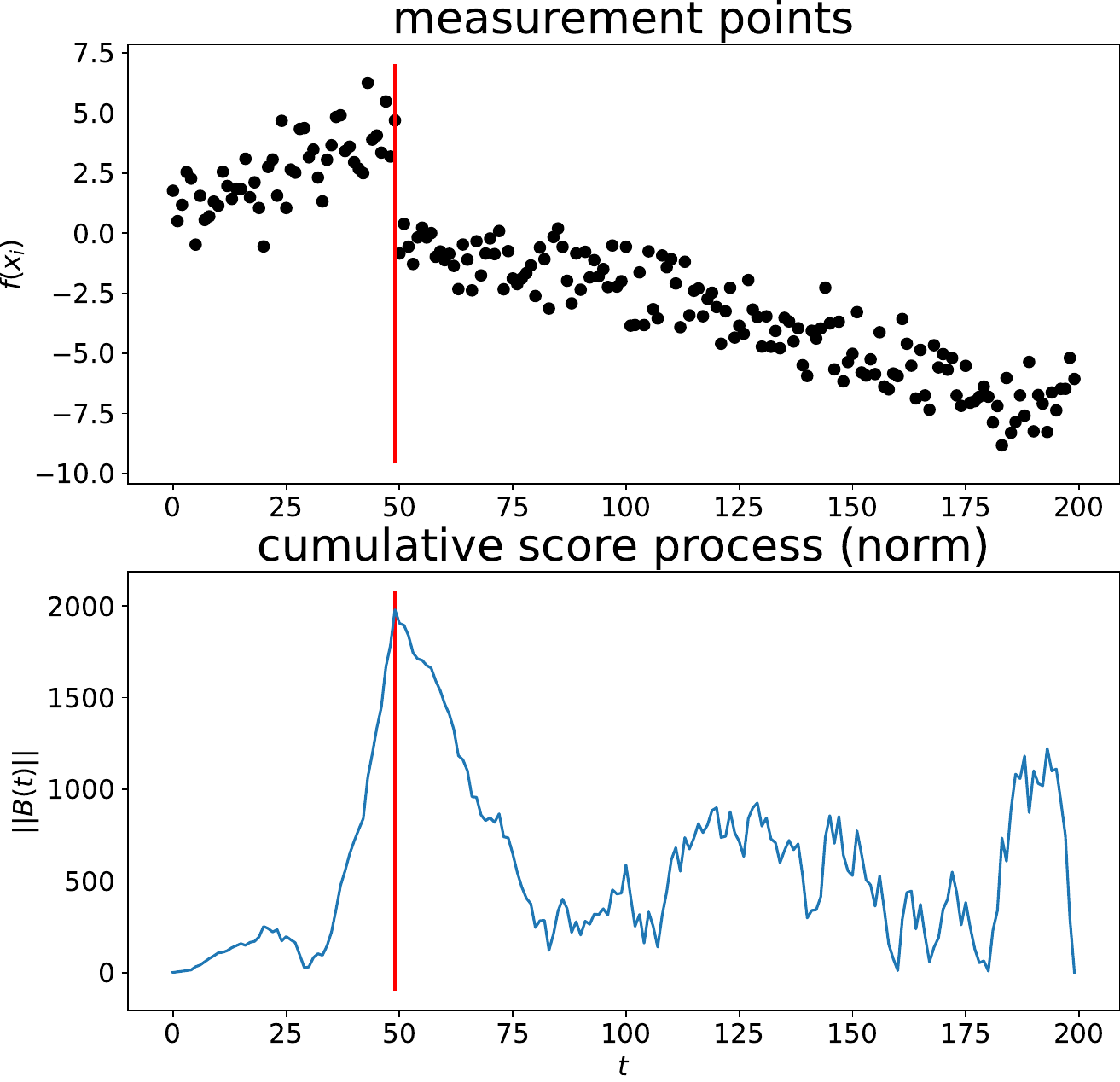}
    \caption{\label{fig:cumulative_process_evolution}Evolution of the norm of the cumulative score process. \emph{Top panel:} measurements of a piecewise-linear model (dark dots) visualized along one axis. \emph{Bottom panel:} evolution of $\norm{B(t)}_1$ as a function of $t$ (solid blue line). The process presents a clear maximum, which gives us a candidate split for this axis (vertical red line).}
\end{figure}

\noindent 
A straightforward computation yields
\begin{equation}
\label{eq:score-specific}
\score{\beta;x_i} = \left( y_i\cdot \xtilde_i - \xtilde_i\xtilde_i^\top \beta \right) \sigma^{-2} 
\, .
\end{equation}
The maximum likelihood estimate $\betahat$ is obtained by solving $\sum_i\score{\beta;x_i} = 0$: in this case this coincides with the \emph{ordinary least-squares} estimate, in a sense the best linear fit on the entire leaf. 
However, what we want is to split this leaf if the deviations from the linear model computed on the whole leaf are too strong. 
Therefore, we compute the \emph{cumulative score process} defined by
\begin{equation}
\label{eq:cumulative-score-process}
\forall t\in [0,1],\quad B^{(j)}(t;\betahat) = \frac{1}{\sqrt{n}} \sum_{i=1}^{\floor{nt}} \score{\betahat ; x_i}
\, ,
\end{equation}
and take as a criterion \textbf{the maximum of the $L^1$ norm of $B^{(j)}(t)$, across all features.} 
The computation of the best split is described in Algorithm~\ref{algo:get_best_split} and we provide an example in Figure~\ref{fig:cumulative_process_evolution}.

Note that Eq.~\eqref{eq:cumulative-score-process} corresponds to Eq.~(13) in \citet{merkle_zeileis_2013}, with the notable difference that Eq.~(13) is normalized (by the square root of the estimated covariance matrix of the scores). 
We observe empirically that normalizing yields worse results when detecting the splits. 

\begin{algorithm}[ht]
  \caption{\label{algo:get_best_split}\texttt{GetBestSplit}: Get the best split for a given leaf.}
  \begin{algorithmic}[1]
    \Require{$x_1,\ldots,x_n$: Sobol points in current leaf; $y_1,\ldots,y_n$: corresponding $\Model$ values}
    \For{$j=1$ to $d$}
\State{re-order $x_1,\ldots,\ldots,x_n$ according to feature $j$}
\State{re-order $y_1,\ldots,y_n$ accordingly}
\State{compute $\betahat=\text{OLS}(y;x)$}
\State{compute predictions $\yhat_i$}
\State{set $\sigma^2=\frac{1}{n}\sum_{i=1}^n (y_i - \yhat_i)^2$}
\State{compute scores $\score{\betahat;x_i}$ according to Eq.~\eqref{eq:score-specific}}
\State{compute $B^{(j)}$ according to Eq.~\eqref{eq:cumulative-score-process}}
\State{store $m_j=\max_{t\in [0,1]} \norm{B^{(j)}(t)}_1$}
\State{store $t_j=\argmax_{t\in [0,1]} \norm{B^{(j)}(t)}_1$}
    \EndFor 
    \State{\Return $\jstar=\argmax_{j\in [d]} m_j$, feature to split}
    \State{\Return $t_\jstar$, value of the split}
    \State{\Return{$\betahat$, local linear model}}
  \end{algorithmic}
\end{algorithm}

\begin{algorithm}[ht]
  \caption{\label{algo:tree-construction}\texttt{RecTree}: Recursive construction of $\surrogatemodel$.}
  \begin{algorithmic}[1]
    \Require{$R$: hyper-rectangle; $S$: Sobol points inside $R$; $V$: values of $\Model$ on $S$}
\State{\textbf{initialize} $T\gets \inputspace$, tree storing partition and models}
\For{ $L\in T.\text{leaves}$}
\If{$R^2(L) \leq \rho$ and $n(L)\geq 2\nmin$}
\State{set $x=\{x_1,\ldots,x_n\} = S\cap L$ }
\State{set $y=\{y_1,\ldots,y_n\}$ accordingly}
\State{$(j,t_j) = $ \texttt{GetBestSplit}$(x,y)$} 
\State{$L^\ell\gets L\cap \{x : x_j \leq t_j\}$}
\State{$L^r\gets L\cap \{x : x_j > t_j\}$}
\State{$L.\text{left\_child}\gets $ \texttt{RecTree}$(L^\ell,S,V)$}
\State{$L.\text{right\_child}$ $\gets$ \texttt{RecTree}$(L^r,S,V)$}
\EndIf 
\EndFor
\Return{$T$}
  \end{algorithmic}
\end{algorithm}

\begin{algorithm}[ht]
  \caption{\label{algo:global}\texttt{GlobalSurrogate}: Compute the global surrogate model used by GLEAMS.}
  \begin{algorithmic}[1]
  \Require{$\Model$: black-box model; $\inputspace$ or its boundaries; $N$: number of Sobol points; $\rho$: $R^2$ threshold; $\nmin$: min. number of points per leaf}
  \State{generate $S=\{x_1,\ldots,x_N\}$ Sobol points in $\inputspace$}
    \State{compute $y_i=\model{x_i}$}
\State{$V\gets \{y_1,\ldots,y_N\}$}
\State{\Return{$T\gets$ \texttt{RecTree}$(\inputspace,S,V)$}}
  \end{algorithmic}
\end{algorithm}

%%%%%%%%%%%%%%%%%%%%%%%%%%%%%%%%%%%%%%%%%%%%%%%%%%%%%%%%%%%%%%%%%%%%%%%

%%%%%%%%%%%%%%%%%%%%%%%%%%%%%%%%%%%%%%%%%%%%%%%%%%%%%%%%%%%%%%%%%%%%%%%%%%%%

\section{Evaluation}
\label{sec:evaluation}

Evaluating interpretability methods is quite challenging: for a given machine learning model, there is rarely a ground-truth available, telling us which features were used for a given prediction. 
\citet{zhou_et_al_2021} outline two main evaluation methodologies: \emph{human-centered} evaluations, and \emph{functionality-grounded} (quantitative, note that the terminology was introduced by \citet{doshi-velez_kim_2017}). 
We focus on the latter, since human-centered evaluations are hard to obtain and can be unreliable. 
Among those, we choose to compare GLEAMS to existing methodology using monotonicity and recall. 
Below we briefly describe these two metrics.

\textbf{Monotonicity} is defined as the correlation between the absolute value of the attributions at a given point and the expected loss of the model restricted to the corresponding feature. 
Intuitively, this takes high values if the model is very imprecise when it does not know the feature at hand. 
This is Metric~2.3 in \citet{nguyen_martinez_2020}. 
Following their notation, for each interpretability method to evaluate, for each point in the test set, we compute on the one hand a vector of attributions $\alpha\in\Reals^d$ and on the other hand a vector of expected loss $e\in\Reals^d$. 
Monotonicity is then defined as the Spearman rank correlation \citep{spearman_1904} between~$\abs{\alpha}$ and $e$. 
Formally, for all $j\in [d]$, $e_j$ is defined as 
\begin{equation}
\label{eq:def-monotonicity}
e_j = \int_{a_j}^{b_j} \ell(\model{\xi},\Model_j(x))p(x)\diff x
\, ,
\end{equation}
where $\Model_j$ is the restriction of $\Model$ to coordinate $j$ (keeping all other coordinates equal to those of $\xi$), $p$ is a probability distribution on $[a_j,b_j]$ (recall that $a_j$ and $b_j$ denote the boundaries of $\inputspace$ along dimension $j$, see Assumption~\ref{ass:rectangle}), and $\ell$ is the squared loss. 

Let us define $\uniform{C}$ the uniform distribution on a compact set $C$. 
In our experiments, we consider two natural choices of $p$ in Eq.~\eqref{eq:def-monotonicity}. 
First, $p\sim \uniform{[\ell_j,r_j]}$, where $\ell_j$ (resp. $r_j$) are the left (resp. right) boundaries of $R$ along feature $j$, where $R$ is the cell containing $\xi$. 
This corresponds to \textbf{local monotonicity}: how well the attributions reflect the variations in prediction locally, that is, from the point of view of GLEAMS, on $R$. 
Second, $p\sim \uniform{[a_j,b_j]}$, which corresponds to \textbf{global monotonicity}: how well the attributions reflect the variations in prediction on the whole input space along feature $j$. 

Let us conclude this section by defining formally the \textbf{Recall of Important Features}. 
For certain models, we can actually be certain that some feature attributions should be~$0$. 
This is the case, for instance, if we force our model $\Model$ to use only a subset of the features. 
As a measure of the quality of explanations, one can then compare the top features selected by an attribution method to the set of true features, which we know. 
Let us call $T$ the set of true features. 
For any given example $\xi$, we can define $T^\xi$ the set of the top $\abs{T}$ features, ranked by $\abs{\alpha_j}$. 
Originally proposed by \citet{ribeiro_et_al_2016}, the recall of important features is then defined as 
\begin{equation}
\label{eq:def-recall}
\rho(\xi) \defeq \frac{\abs{T\cap T^\xi}}{\abs{T}}
\, .
\end{equation}
Intuitively, $\rho(\xi)$ is large (close to one) if the considered attribution method gives large attributions to the true features. 
In our experiments, we report the average recall over all points of the test set unless otherwise mentioned. 

%%%%%%%%%%%%%%%%%%%%%%%%%%%%%%%%%%%%%%%%%%%%%%%%%%%%%%%%%%%%%%%%%%%%%%%%%%%

\section{Toy data experiments}
\label{sec:toy-data}

%%%%%%%%%%%%%%%%%%%%%%%%%%%%%%%%%%%%%%%%%

\begin{figure*}[ht!]
\centering
\includegraphics[scale=0.32]{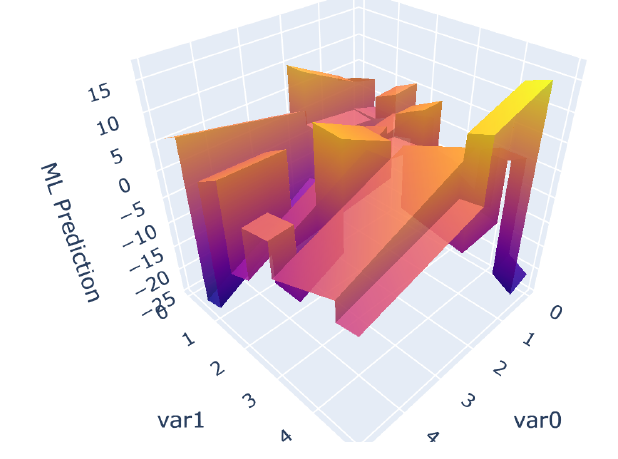}
\hfill 
\includegraphics[scale=0.32]{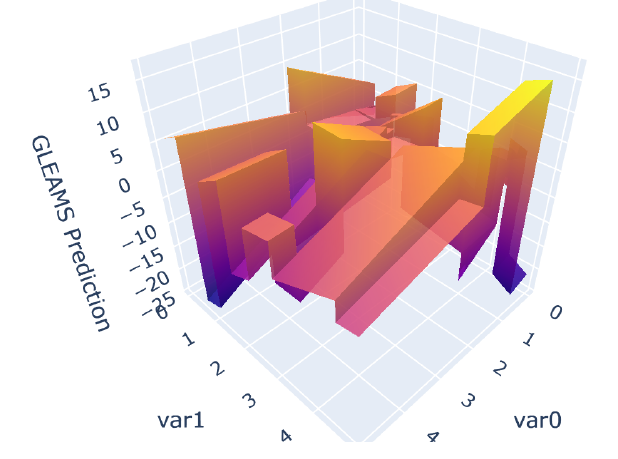}
\hfill 
\includegraphics[scale=0.32]{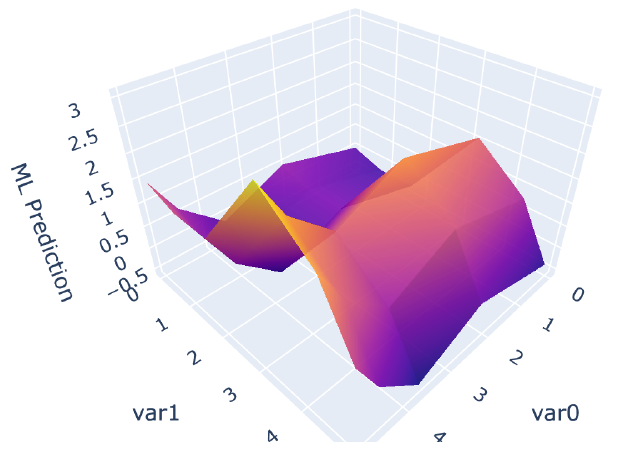}
\hfill 
\includegraphics[scale=0.32]{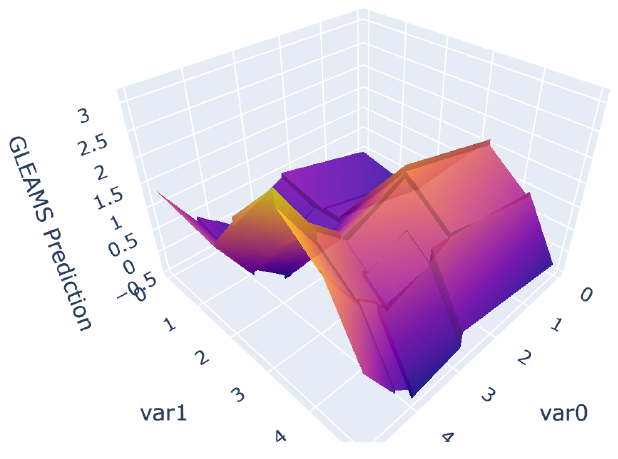}
\caption{\label{fig:toy-sanity-check}Approximating piecewise-linear models with GLEAMS surrogate model. \emph{Left panels:} example (i), model on the left and GLEAMS surrogate on the right. \emph{Right panels:} example (ii), model on the left and GLEAMS surrogate on the right.}
\end{figure*}

%%%%%%%%%%%%%%%%%%%%%%%%%%%%%%%%%%%%%%%%%%%%%%%%%%%%%%%%%%

As a first sanity check, we ran the partition scheme of GLEAMS on simple models and  checked whether the surrogate model~$\surrogatemodel$ was indeed close to the true model~$\Model$. 
In order to have a ground-truth for the partition, we chose partition-based models, \emph{i.e.}, there exist a partition $\partition$ of $\inputspace$ such that the parameters of~$\Model$ are constant on each rectangular cell $R\in\partition$. 
We considered a simple setting where the model is piecewise-linear with rectangular cells,  a situation in which GLEAMS could, in theory, recover perfectly both the underlying partition and the coefficients of the local models. 
We present here two examples: (i) a discontinuous piecewise-linear model with arbitrary coefficients, and (ii) a piecewise-linear model with continuity constraints. 
Example (i) is a generalization of the surface produced by a Random Forest regressor, while example (ii) is reminiscent of a fully-connected, feedforward, ReLU activated neural network. 
Note however that for the latter, the cells of the partition are more complicated than simple rectangle, see \citet{montufar_et_al_2014} (in particular Figure~2). 
In each case, GLEAMS recovers perfectly the underlying model, as demonstrated in Figure~\ref{fig:gleams-explanations}.

It is interesting to notice that, despite having a stopping criterion on the coefficient of determination, not all leaves satisfy this criterion since the other stopping criterion is a minimal number of points per leaf. 
Thus it is possible to end up in situations where the $R^2$ on a given leaf is sub par. 
Nevertheless, these leaves are by construction quite small. 
For instance, in example (i), the average $R^2$ is equal to $1$, and in example (ii) it is equal to $.98$.

\end{document}

%% file: macros.tex
\theoremstyle{plain}
\theoremstyle{definition}
\newtheorem{assumption}{Assumption}%[section]
\DeclareMathOperator*{\argmax}{arg\max}

\newcommand{\abs}[1]{\left\lvert#1\right\rvert}% absolute value
\newcommand{\defeq}{\vcentcolon =}% define equals
\newcommand{\diff}{\mathrm{d}}% diff operator
\newcommand{\expl}[1]{\mathrm{exp}\left(#1\right)}% exp of something
\newcommand{\floor}[1]{\left\lfloor #1 \right\rfloor}% floor function
\newcommand{\norm}[1]{\left\lVert #1\right\rVert}% norm of something
\newcommand{\Reals}{\mathbb{R}}% real numbers
\newcommand{\Uniform}{\mathcal{U}}% uniform distribution
\newcommand{\uniform}[1]{\Uniform(#1)}% uniform distribution on a compact
\newcommand{\volume}[1]{\mathcal{V}(#1)}% volume of something

\newcommand{\betahat}{\hat{\beta}}% estimate of \beta
\newcommand{\inputspace}{\mathcal{X}}% input space
\newcommand{\jstar}{j^\star}% best feature
\newcommand{\labels}{\mathcal{Y}}% output space
\newcommand{\Likelihood}{\mathcal{L}}% likelihood function
\newcommand{\likelihood}[1]{\Likelihood(#1)}% likelihood of something
\newcommand{\Model}{f}% symbol for the black-box model to explain
\newcommand{\model}[1]{\Model(#1)}% model applied to something
\newcommand{\nmin}{n_\text{min}}% minimal number of points in the leaf
\newcommand{\partition}{\mathcal{P}}% partition of the input space
\newcommand{\Score}{s}% scores
\newcommand{\score}[1]{\Score(#1)}% score of something
\newcommand{\surrogatemodel}{\hat{f}}% global surrogate
\newcommand{\xtilde}{\tilde{x}}% x with padding
\newcommand{\yhat}{\hat{y}}% predictions

%% file: paper.bbl
\begin{thebibliography}{29}
\expandafter\ifx\csname natexlab\endcsname\relax\def\natexlab#1{#1}\fi
\providecommand{\url}[1]{\texttt{#1}}
\providecommand{\href}[2]{#2}
\providecommand{\path}[1]{#1}
\providecommand{\DOIprefix}{doi:}
\providecommand{\ArXivprefix}{arXiv:}
\providecommand{\URLprefix}{URL: }
\providecommand{\Pubmedprefix}{pmid:}
\providecommand{\doi}[1]{\href{http://dx.doi.org/#1}{\path{#1}}}
\providecommand{\Pubmed}[1]{\href{pmid:#1}{\path{#1}}}
\providecommand{\bibinfo}[2]{#2}
\ifx\xfnm\relax \def\xfnm[#1]{\unskip,\space#1}\fi
%Type = Article
\bibitem[{Borisov et~al.(2022)Borisov, Leemann, Se{\ss}ler, Haug, Pawelczyk,
  and Kasneci}]{borisov_et_al_2021}
\bibinfo{author}{V.~Borisov}, \bibinfo{author}{T.~Leemann},
  \bibinfo{author}{K.~Se{\ss}ler}, \bibinfo{author}{J.~Haug},
  \bibinfo{author}{M.~Pawelczyk}, \bibinfo{author}{G.~Kasneci},
\newblock \bibinfo{title}{Deep neural networks and tabular data: A survey},
\newblock \bibinfo{journal}{IEEE Transactions on Neural Networks and Learning
  Systems}  (\bibinfo{year}{2022}).
%Type = Article
\bibitem[{Wachter et~al.(2017)Wachter, Mittelstadt, and
  Floridi}]{wachter_et_al_2017}
\bibinfo{author}{S.~Wachter}, \bibinfo{author}{B.~Mittelstadt},
  \bibinfo{author}{L.~Floridi},
\newblock \bibinfo{title}{Why a {{Right}} to {{Explanation}} of {{Automated
  Decision-Making Does Not Exist}} in the {{General Data Protection
  Regulation}}},
\newblock \bibinfo{journal}{International Data Privacy Law} \bibinfo{volume}{7}
  (\bibinfo{year}{2017}) \bibinfo{pages}{76--99}.
%Type = Article
\bibitem[{Adadi and Berrada(2018)}]{adadi_berrada_2018}
\bibinfo{author}{A.~Adadi}, \bibinfo{author}{M.~Berrada},
\newblock \bibinfo{title}{Peeking {{Inside}} the {{Black-Box}}: {{A Survey}} on
  {{Explainable Artificial Intelligence}} ({{XAI}})},
\newblock \bibinfo{journal}{IEEE Access} \bibinfo{volume}{6}
  (\bibinfo{year}{2018}) \bibinfo{pages}{52138--52160}.
%Type = Article
\bibitem[{Linardatos et~al.(2020)Linardatos, Papastefanopoulos, and
  Kotsiantis}]{linardatos_et_al_2020}
\bibinfo{author}{P.~Linardatos}, \bibinfo{author}{V.~Papastefanopoulos},
  \bibinfo{author}{S.~Kotsiantis},
\newblock \bibinfo{title}{{Explainable AI: A review of machine learning
  interpretability methods}},
\newblock \bibinfo{journal}{Entropy} \bibinfo{volume}{23}
  (\bibinfo{year}{2020}) \bibinfo{pages}{18}.
%Type = Inproceedings
\bibitem[{Ribeiro et~al.(2016)Ribeiro, Singh, and
  Guestrin}]{ribeiro_et_al_2016}
\bibinfo{author}{M.~T. Ribeiro}, \bibinfo{author}{S.~Singh},
  \bibinfo{author}{C.~Guestrin},
\newblock \bibinfo{title}{{Why Should I Trust You?: Explaining the Predictions
  of Any Classifier}},
\newblock in: \bibinfo{booktitle}{Proceedings of the 22nd {{ACM SIGKDD}}
  International Conference on Knowledge Discovery and Data Mining},
  \bibinfo{publisher}{{ACM}}, \bibinfo{year}{2016}, pp.
  \bibinfo{pages}{1135--1144}.
%Type = Article
\bibitem[{Sobol(1985)}]{sobol_1985}
\bibinfo{author}{I.~M. Sobol},
\newblock \bibinfo{title}{Points which uniformly fill a multidimensional cube},
\newblock \bibinfo{journal}{Mathematics, cybernetics series}
  (\bibinfo{year}{1985}) \bibinfo{pages}{32}.
%Type = Book
\bibitem[{Breiman et~al.(1984)Breiman, Friedman, Olshen, and
  Stone}]{breiman_et_al_1984}
\bibinfo{author}{L.~Breiman}, \bibinfo{author}{J.~H. Friedman},
  \bibinfo{author}{R.~A. Olshen}, \bibinfo{author}{C.~J. Stone},
  \bibinfo{title}{Classification and regression trees},
  \bibinfo{publisher}{Wadsworth \& Brooks / Cole Advanced Books \& Software},
  \bibinfo{year}{1984}.
%Type = Article
\bibitem[{Chan and Loh(2004)}]{chan_loh_2004}
\bibinfo{author}{K.-Y. Chan}, \bibinfo{author}{W.-Y. Loh},
\newblock \bibinfo{title}{{LOTUS: An algorithm for building accurate and
  comprehensible logistic regression trees}},
\newblock \bibinfo{journal}{Journal of Computational and Graphical Statistics}
  \bibinfo{volume}{13} (\bibinfo{year}{2004}) \bibinfo{pages}{826--852}.
%Type = Article
\bibitem[{Zeileis et~al.(2008)Zeileis, Hothorn, and
  Hornik}]{zeileis_et_al_2008}
\bibinfo{author}{A.~Zeileis}, \bibinfo{author}{T.~Hothorn},
  \bibinfo{author}{K.~Hornik},
\newblock \bibinfo{title}{Model-based recursive partitioning},
\newblock \bibinfo{journal}{Journal of Computational and Graphical Statistics}
  \bibinfo{volume}{17} (\bibinfo{year}{2008}) \bibinfo{pages}{492--514}.
%Type = Article
\bibitem[{Cortez et~al.(2009)Cortez, Cerdeira, Almeida, Matos, and
  Reis}]{cortez_et_al_2009}
\bibinfo{author}{P.~Cortez}, \bibinfo{author}{A.~Cerdeira},
  \bibinfo{author}{F.~Almeida}, \bibinfo{author}{T.~Matos},
  \bibinfo{author}{J.~Reis},
\newblock \bibinfo{title}{Modeling wine preferences by data mining from
  physicochemical properties},
\newblock \bibinfo{journal}{Decision support systems} \bibinfo{volume}{47}
  (\bibinfo{year}{2009}) \bibinfo{pages}{547--553}.
%Type = Article
\bibitem[{Tsanas et~al.(2009)Tsanas, Little, McSharry, and
  Ramig}]{tsanas_et_al_2009}
\bibinfo{author}{A.~Tsanas}, \bibinfo{author}{M.~Little},
  \bibinfo{author}{P.~McSharry}, \bibinfo{author}{L.~Ramig},
\newblock \bibinfo{title}{Accurate telemonitoring of parkinson’s disease
  progression by non-invasive speech tests},
\newblock \bibinfo{journal}{Nature Precedings}  (\bibinfo{year}{2009})
  \bibinfo{pages}{1--1}.
%Type = Inproceedings
\bibitem[{Chen and Guestrin(2016)}]{chen_guestrin_2016}
\bibinfo{author}{T.~Chen}, \bibinfo{author}{C.~Guestrin},
\newblock \bibinfo{title}{{{XGBoost}}: {{A Scalable Tree Boosting System}}},
\newblock in: \bibinfo{booktitle}{Proceedings of the 22nd {{ACM SIGKDD
  International Conference}} on {{Knowledge Discovery}} and {{Data Mining}}},
  \bibinfo{year}{2016}, pp. \bibinfo{pages}{785--794}.
%Type = Article
\bibitem[{Friedman(2001)}]{friedman2001greedy}
\bibinfo{author}{J.~H. Friedman},
\newblock \bibinfo{title}{Greedy function approximation: a gradient boosting
  machine},
\newblock \bibinfo{journal}{The Annals of Statistics}  (\bibinfo{year}{2001})
  \bibinfo{pages}{1189--1232}.
%Type = Article
\bibitem[{Kingma and Ba(2014)}]{kingma_ba_2014}
\bibinfo{author}{D.~P. Kingma}, \bibinfo{author}{J.~Ba},
\newblock \bibinfo{title}{{Adam: A method for stochastic optimization}},
\newblock \bibinfo{journal}{arXiv preprint arXiv:1412.6980}
  (\bibinfo{year}{2014}).
%Type = Article
\bibitem[{Nguyen and Mart{\'\i}nez(2020)}]{nguyen_martinez_2020}
\bibinfo{author}{A.-P. Nguyen}, \bibinfo{author}{M.~R. Mart{\'\i}nez},
\newblock \bibinfo{title}{On quantitative aspects of model interpretability},
\newblock \bibinfo{journal}{arXiv preprint arXiv:2007.07584}
  (\bibinfo{year}{2020}).
%Type = Article
\bibitem[{Craven and Shavlik(1996)}]{craven_shavlik_1996}
\bibinfo{author}{M.~Craven}, \bibinfo{author}{J.~W. Shavlik},
\newblock \bibinfo{title}{Extracting tree-structured representations of trained
  networks},
\newblock \bibinfo{journal}{Advances in Neural Information Processing Systems}
  (\bibinfo{year}{1996}) \bibinfo{pages}{24--30}.
%Type = Book
\bibitem[{Quinlan(1993)}]{quinlan_1993}
\bibinfo{author}{J.~R. Quinlan}, \bibinfo{title}{{C4.5: Programs for Machine
  Learning}}, \bibinfo{publisher}{Elsevier}, \bibinfo{year}{1993}.
%Type = Article
\bibitem[{Gibbons et~al.(2013)Gibbons, Hooker, Finkelman, Weiss, Pilkonis,
  Frank, Moore, and Kupfer}]{gibbons_et_al_2013}
\bibinfo{author}{R.~D. Gibbons}, \bibinfo{author}{G.~Hooker},
  \bibinfo{author}{M.~D. Finkelman}, \bibinfo{author}{D.~J. Weiss},
  \bibinfo{author}{P.~A. Pilkonis}, \bibinfo{author}{E.~Frank},
  \bibinfo{author}{T.~Moore}, \bibinfo{author}{D.~J. Kupfer},
\newblock \bibinfo{title}{The {{CAD-MDD}}: A computerized adaptive diagnostic
  screening tool for depression},
\newblock \bibinfo{journal}{The Journal of clinical psychiatry}
  \bibinfo{volume}{74} (\bibinfo{year}{2013}) \bibinfo{pages}{669}.
%Type = Article
\bibitem[{Zhou and Hooker(2016)}]{zhou_hooker_2016}
\bibinfo{author}{Y.~Zhou}, \bibinfo{author}{G.~Hooker},
\newblock \bibinfo{title}{Interpreting {{Models}} via {{Single Tree
  Approximation}}}  (\bibinfo{year}{2016}).
%Type = Article
\bibitem[{Lei et~al.(2018)Lei, G'Sell, Rinaldo, Tibshirani, and
  Wasserman}]{lei_distribution-free_2018}
\bibinfo{author}{J.~Lei}, \bibinfo{author}{M.~G'Sell},
  \bibinfo{author}{A.~Rinaldo}, \bibinfo{author}{R.~J. Tibshirani},
  \bibinfo{author}{L.~Wasserman},
\newblock \bibinfo{title}{Distribution-free predictive inference for
  regression},
\newblock \bibinfo{journal}{Journal of the American Statistical Association}
  \bibinfo{volume}{113} (\bibinfo{year}{2018}) \bibinfo{pages}{1094--1111}.
%Type = Article
\bibitem[{Lundberg and Lee(2017)}]{lundberg2017unified}
\bibinfo{author}{S.~M. Lundberg}, \bibinfo{author}{S.-I. Lee},
\newblock \bibinfo{title}{A unified approach to interpreting model
  predictions},
\newblock \bibinfo{journal}{Advances in neural information processing systems}
  \bibinfo{volume}{30} (\bibinfo{year}{2017}).
%Type = Inproceedings
\bibitem[{Ribeiro et~al.(2018)Ribeiro, Singh, and
  Guestrin}]{ribeiro_et_al_2018}
\bibinfo{author}{M.~T. Ribeiro}, \bibinfo{author}{S.~Singh},
  \bibinfo{author}{C.~Guestrin},
\newblock \bibinfo{title}{Anchors: {{High-precision}} model-agnostic
  explanations},
\newblock in: \bibinfo{booktitle}{Thirty-{{Second AAAI Conference}} on
  {{Artificial Intelligence}}}, \bibinfo{year}{2018}.
%Type = Article
\bibitem[{Setzu et~al.(2021)Setzu, Guidotti, Monreale, Turini, Pedreschi, and
  Giannotti}]{setzu_et_al_2021}
\bibinfo{author}{M.~Setzu}, \bibinfo{author}{R.~Guidotti},
  \bibinfo{author}{A.~Monreale}, \bibinfo{author}{F.~Turini},
  \bibinfo{author}{D.~Pedreschi}, \bibinfo{author}{F.~Giannotti},
\newblock \bibinfo{title}{{Glocalx-from local to global explanations of black
  box AI models}},
\newblock \bibinfo{journal}{Artificial Intelligence} \bibinfo{volume}{294}
  (\bibinfo{year}{2021}) \bibinfo{pages}{103457}.
%Type = Inproceedings
\bibitem[{Harder et~al.(2020)Harder, Bauer, and Park}]{harder_et_al_2020}
\bibinfo{author}{F.~Harder}, \bibinfo{author}{M.~Bauer},
  \bibinfo{author}{M.~Park},
\newblock \bibinfo{title}{Interpretable and differentially private
  predictions},
\newblock in: \bibinfo{booktitle}{Proceedings of the AAAI Conference on
  Artificial Intelligence}, volume~\bibinfo{volume}{34}, \bibinfo{year}{2020},
  pp. \bibinfo{pages}{4083--4090}.
%Type = Article
\bibitem[{Merkle and Zeileis(2013)}]{merkle_zeileis_2013}
\bibinfo{author}{E.~C. Merkle}, \bibinfo{author}{A.~Zeileis},
\newblock \bibinfo{title}{{Tests of Measurement Invariance without Subgroups: A
  Generalization of Classical Methods}},
\newblock \bibinfo{journal}{Psychometrika} \bibinfo{volume}{78}
  (\bibinfo{year}{2013}) \bibinfo{pages}{59--82}.
%Type = Article
\bibitem[{Zhou et~al.(2021)Zhou, Gandomi, Chen, and
  Holzinger}]{zhou_et_al_2021}
\bibinfo{author}{J.~Zhou}, \bibinfo{author}{A.~H. Gandomi},
  \bibinfo{author}{F.~Chen}, \bibinfo{author}{A.~Holzinger},
\newblock \bibinfo{title}{{Evaluating the quality of machine learning
  explanations: A survey on methods and metrics}},
\newblock \bibinfo{journal}{Electronics} \bibinfo{volume}{10}
  (\bibinfo{year}{2021}) \bibinfo{pages}{593}.
%Type = Article
\bibitem[{Doshi-Velez and Kim(2017)}]{doshi-velez_kim_2017}
\bibinfo{author}{F.~Doshi-Velez}, \bibinfo{author}{B.~Kim},
\newblock \bibinfo{title}{Towards a rigorous science of interpretable machine
  learning},
\newblock \bibinfo{journal}{arXiv preprint arXiv:1702.08608}
  (\bibinfo{year}{2017}).
%Type = Article
\bibitem[{Spearman(1904)}]{spearman_1904}
\bibinfo{author}{C.~Spearman},
\newblock \bibinfo{title}{The proof and measurement of association between two
  things},
\newblock \bibinfo{journal}{The American journal of psychology}
  \bibinfo{volume}{100} (\bibinfo{year}{1904}) \bibinfo{pages}{441--471}.
%Type = Article
\bibitem[{Montufar et~al.(2014)Montufar, Pascanu, Cho, and
  Bengio}]{montufar_et_al_2014}
\bibinfo{author}{G.~F. Montufar}, \bibinfo{author}{R.~Pascanu},
  \bibinfo{author}{K.~Cho}, \bibinfo{author}{Y.~Bengio},
\newblock \bibinfo{title}{On the number of linear regions of deep neural
  networks},
\newblock \bibinfo{journal}{Advances in neural information processing systems}
  \bibinfo{volume}{27} (\bibinfo{year}{2014}).

\end{thebibliography}
